\documentclass{article}

\usepackage{PRIMEarxiv}

\usepackage[utf8]{inputenc} 
\usepackage[T1]{fontenc}    
\usepackage{hyperref}       
\usepackage{url}            
\usepackage{amsfonts}       
\usepackage{nicefrac}       
\usepackage{microtype}      
\usepackage{lipsum}
\usepackage{graphicx}
\graphicspath{ {./images/} }
\usepackage{float}
\usepackage{tabularx}
\usepackage{threeparttable}
\usepackage{multirow}
\usepackage{enumitem}
\usepackage{cite}

\usepackage[table]{xcolor}
\usepackage{booktabs} 
\usepackage{arydshln}
\usepackage[normalem]{ulem}
\usepackage{pifont}
\definecolor{lightred}{HTML}{FEC8C8}
\definecolor{lightblue}{HTML}{C8DCFF}
\newcommand{\best}[1]{\uline{\textbf{#1}}}

\newcommand{\colorboxrule}[2]{%
  \begingroup
  \color{#1}
  \rule{6pt}{6pt}
  \endgroup
  #2%
}

\title{TCM-5CEval: Extended Deep Evaluation Benchmark for LLM’s Comprehensive Clinical Research Competence in Traditional Chinese Medicine}
\author{
  Tianai Huang\textsuperscript{1,†},
  Jiayuan Chen\textsuperscript{2,†},
  Lu Lu\textsuperscript{2},
  Pengcheng Chen\textsuperscript{3},
  Tianbin Li\textsuperscript{2},
  Bing Han\textsuperscript{2},
  Wenchao Tang\textsuperscript{1}*,\\
  \textbf{Jie Xu\textsuperscript{2}*,
  Ming Li\textsuperscript{1}*}  \\
  \vspace{0.5em}\\
  \textsuperscript{1}School of Artificial Intelligence in Traditional Chinese Medicine, Shanghai University of Traditional Chinese \\Medicine, Shanghai, China \\
  \textsuperscript{2}Shanghai Artificial Intelligence Laboratory, Shanghai, China \\
   \textsuperscript{3}University of Washington, Seattle, Washington, US \\
   \vspace{0.5em}\\
  *Correspondence to: \texttt{\{Xu Jie\}xujie@pjlab.org.cn}, \texttt{\{Tang Wenchao\}vincent.tang@shutcm.edu.cn},\\ \texttt{\{Li Ming \}acupunture@126.com}\\
  †These authors contributed equally.
}

\begin{document}
\maketitle
\begin{abstract}
 Large language models (LLMs) have demonstrated exceptional capabilities in general domains, yet their application in highly specialized and culturally-rich fields like Traditional Chinese Medicine (TCM) requires rigorous and nuanced evaluation. Building upon prior foundational work such as TCM-3CEval, which highlighted systemic knowledge gaps and the importance of cultural-contextual alignment, we introduce TCM-5CEval, a more granular and comprehensive benchmark. TCM-5CEval is designed to assess LLMs across five critical dimensions: \textbf{(1) Core Knowledge (TCM-Exam)}, \textbf{(2) Classical Literacy (TCM-LitQA)}, \textbf{(3) Clinical Decision-making (TCM-MRCD)}, \textbf{(4) Chinese Materia Medica (TCM-CMM)}, and \textbf{(5) Clinical Non-pharmacological Therapy (TCM-ClinNPT)}. We conducted a thorough evaluation of fifteen prominent LLMs, revealing significant performance disparities and identifying top-performing models like deepseek\_r1 and gemini\_2\_5\_pro. Our findings show that while models exhibit proficiency in recalling foundational knowledge, they struggle with the interpretative complexities of classical texts. Critically, permutation-based consistency testing reveals widespread fragilities in model inference. All evaluated models, including the highest-scoring ones, displayed a substantial performance degradation when faced with varied question option ordering, indicating a pervasive sensitivity to positional bias and a lack of robust understanding. TCM-5CEval not only provides a more detailed diagnostic tool for LLM capabilities in TCM but aldso exposes fundamental weaknesses in their reasoning stability. To promote further research and standardized comparison, TCM-5CEval has been uploaded to the Medbench platform, joining its predecessor in the "In-depth Challenge for Comprehensive TCM Abilities" special track.
\end{abstract}


\section{Introduction}
In recent years, Large Language Models (LLMs) have achieved remarkable progress in natural language processing, with their applications expanding into specialized medical domains\cite{shool2025systematic}. However, when it comes to Traditional Chinese Medicine (TCM) - a field characterized by its unique theoretical system and diagnostic features - existing evaluation methods face considerable challenges\cite{chen2024traditional}. Rooted in traditional Chinese culture, TCM emphasizes ‘holistic concepts’\cite{dubey2025comprehensive} and ‘treatment based on syndrome differentiation.’\cite{zhang2025exploration} Its knowledge system encompasses abstract notions such as yin-yang and five elements theory, zang-fu organ and meridian systems, along with complex pattern identification, formula composition, and medicinal property classification\cite{leong2024dao}. Current evaluation methods primarily rely on objective question types, which may not adequately assess LLMs' deep reasoning capabilities and logical thinking processes\cite{cheng2025empowering, lee2024reasoning}. This limitation in evaluation methodology makes it difficult to accurately measure LLMs' true proficiency in TCM, particularly in assessing capabilities that require complex diagnostic thinking, such as clinical decision-making and classical literature interpretation. Furthermore, existing evaluation systems show room for improvement in covering non-pharmacological therapies like acupuncture and tuina, as well as in-depth knowledge of Chinese materia medica. These considerations highlight opportunities for enhancing the application of artificial intelligence in TCM preservation and innovation\cite{lu2024ai, li2024opportunities, zhou2024integrating}. Therefore, developing a comprehensive framework that expands evaluation dimensions while innovating assessment methods represents a meaningful direction for supporting TCM's digital advancement.

In the field of TCM LLM evaluation, existing research has primarily focused on developing specialized benchmarks, which can be grouped by their distinct focuses and methodologies. One line of research has centered on standardized knowledge assessment, often leveraging examination content. For instance, TCMBench introduces the TCM-ED dataset, consisting of 5,473 questions derived from the TCM Licensing Examination (TCMLE) to systematically cover core theories and clinical practice \cite{yue2024tcmbench}. Similarly, TCMD focuses on constructing a large-scale medical question-answering dataset with manually designed instructions specifically tailored for TCM examination tasks \cite{yu2024tcmd}. Other frameworks have targeted deeper, more specific clinical reasoning skills or aimed for broader, multidimensional coverage. In the former category, TCMEval-SDT provides a specialized benchmark of 300 real-world cases, sourced from diverse records and meticulously annotated by experts, to specifically assess the complex dialectical thinking processes in syndrome differentiation \cite{wang2025tcmeval}. In the latter category, MTCMB proposes a comprehensive framework that integrates real-world case records, exam content, and classical literature across 12 sub-datasets, covering five major categories: knowledge quality, language understanding, diagnostic reasoning, prescription generation, and safety \cite{kong2025mtcmb}. Our prior work, TCM-3CEval, also follows this multidimensional approach, evaluating models across three dimensions—core knowledge, classical literacy, and clinical decision-making \cite{huang2025tcm}. Finally, some work has pushed into novel modalities, most notably TCM-Ladder, a multimodal QA dataset that incorporates images and videos alongside various question formats to evaluate visual understanding \cite{xie2025tcm}. 

Although these specialized evaluation frameworks have demonstrated certain capabilities in handling professional tasks, their assessments primarily emphasize objective knowledge evaluation, with relatively limited investigation into reasoning processes and cognitive abilities. Research indicates that current evaluation methods perform reasonably well in assessing theoretical knowledge retention \cite{yue2024tcmbench, yu2024tcmd} but appear less suited for evaluating competencies requiring subjective judgment \cite{wang2025tcmeval}. Existing evaluation systems therefore present clear opportunities for development in two key areas: first, in methodological approaches, particularly in implementing effective subjective assessment mechanisms; and second, in dimensional coverage. The scope of evaluation is often limited to clinical practice, with key practical domains under-represented and dimensions related to TCM scientific research almost entirely overlooked. These gaps limit a truly comprehensive understanding of LLMs' TCM capabilities.

This study builds upon the existing 3C evaluation framework to propose an expanded ‘5C’ comprehensive evaluation framework for TCM LLMs. The new framework maintains the three original core dimensions while incorporating two additional dimensions: Clinical Non-pharmacological Therapy and Chinese Materia Medica, forming a more complete architecture for assessing TCM knowledge systems. In terms of methodological development, we have explored integrating both subjective and objective assessment approaches across all dimensions. For Core Knowledge, besides conventional single-choice and multiple-choice questions, open-ended questions are introduced to examine theoretical understanding depth. The Classical Literacy dimension employs text interpretation exercises to assess classical literature comprehension, while Clinical Decision-Making utilizes case analysis to evaluate diagnostic reasoning. Clinical Non-pharmacological Therapy incorporates treatment design tasks to examine therapeutic application skills, and Chinese Materia Medica includes formula analysis to assess medicinal compatibility logic. This combined assessment approach seeks to provide a more comprehensive evaluation of LLMs' TCM capabilities across both knowledge breadth and cognitive depth, potentially addressing some limitations of previous frameworks in dimension coverage while offering complementary perspectives to solely objective-question-based evaluation. We hope these exploratory efforts might contribute to establishing more nuanced evaluation standards for TCM LLM development, while possibly supporting the advancement of TCM digitalization research in more substantive directions.

\section{Methods}

\subsection{Evaluation dimension design}

\subsubsection{Data source}

Based on TCM-3CEVAL, we propose TCM-5CEVAL, a Traditional Chinese Medicine evaluation benchmark that utilizes exercise sets from nationally authorized planning textbooks as data sources, and is deeply aligned with TCM teaching outline and clinical disciplines. The exercise sets of TCM planning textbooks in this study primarily consist of those from the National Higher Education TCM Planning Textbook series published by China Press of Traditional Chinese Medicine, with the main versions being the "13th Five-Year Plan Textbook" and "14th Five-Year Plan Textbook" exercise sets. This textbook series was compiled under the guidance of the National Administration of Traditional Chinese Medicine, with collaboration between China Press of Traditional Chinese Medicine and authoritative experts from various fields across national TCM institutions. Currently, the "13th Five-Year Plan Textbook" and "14th Five-Year Plan Textbook" are widely adopted as authoritative teaching materials in TCM institutions nationwide.

\begin{figure}
    \centering
    \includegraphics[width=1\linewidth]{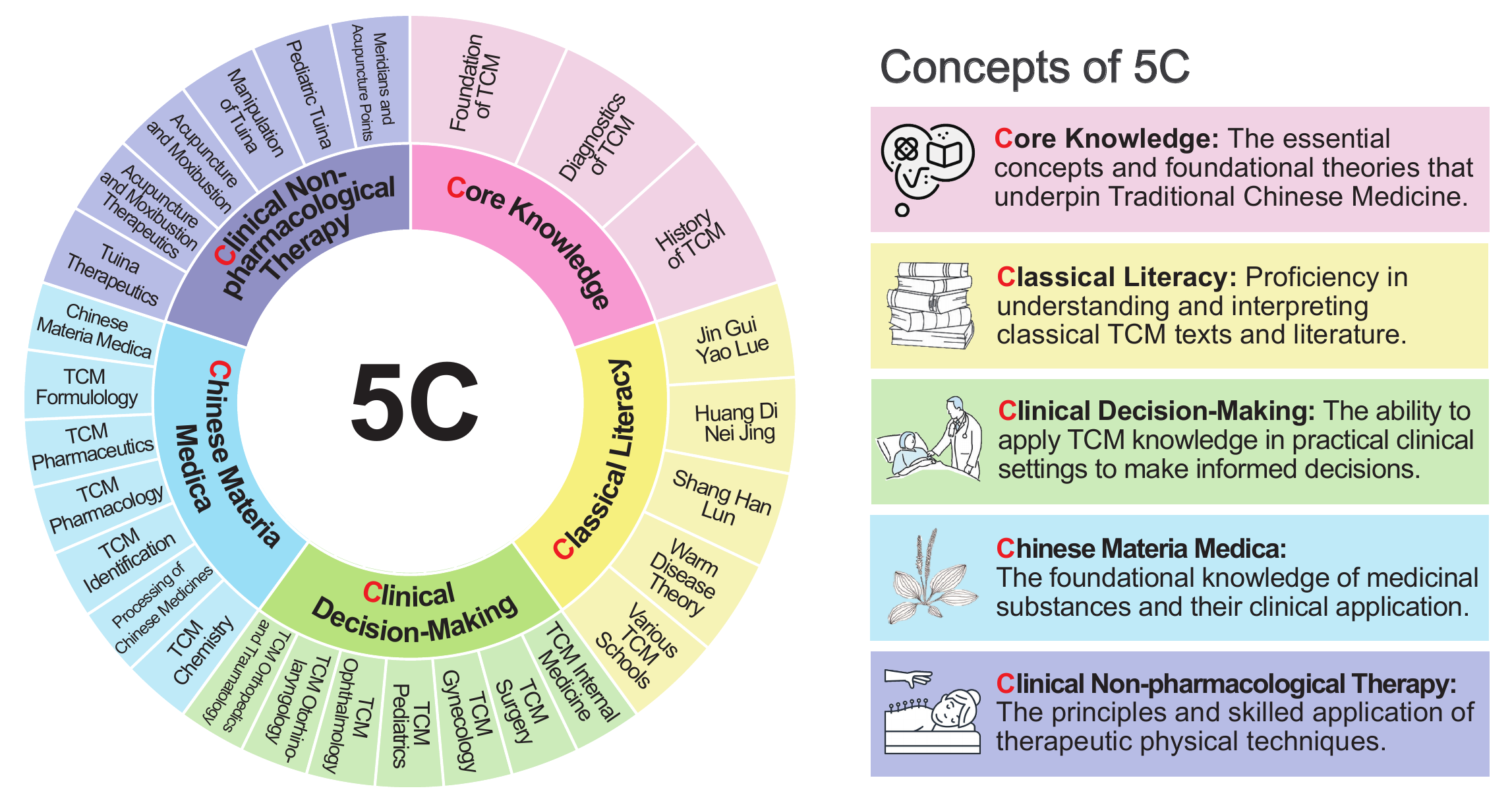}
    \caption{Overview diagram of TCM 5C-EVAL}
    \label{fig:Overview diagram of TCM 5C-EVAL}
\end{figure}

This study selected exercise sets from 30 TCM-related courses, including Chinese Internal Medicine, Chinese External Medicine, Gynecology of TCM, Pediatrics of TCM, Ophthalmology of TCM, Otorhinolaryngology of TCM, Orthopedics and Traumatology of TCM, Basic Theory of TCM, Diagnostics of TCM, Chinese Medicinal Formulas, Theories of Schools of TCM, Treatise on Cold Damage Diseases, Theory and Practice of Chinese Medicines, Acupuncture and Moxibustion, Meridians and Acupuncture Points, and Tuina, as the data sources for TCM-5CEVAL.The 30 TCM-related exercise sets were categorized into the following five modules based on clinical practice and theoretical frameworks: Clinical Decision-making, Core Knowledge, Classical Literacy, Chinese Materia Medica, and Clinical Non-pharmacological Therapy. The specific data sources and classifications are illustrated in \autoref{fig:Overview diagram of TCM 5C-EVAL}.

\subsubsection{Classification criteria}
The classification criteria of this study are as follows:In Ming Dynasty China, medicine was divided into thirteen specialties: general internal medicine, pediatric medicine, gynecology, ulcer treatment, acupuncture and moxibustion, ophthalmology, oral-dental medicine, laryngology, fracture and cold damage treatment, wound treatment, massage, and incantation therapy.

The Undergraduate Program Catalog of Regular Higher Education Institutions (2025) issued by China's Ministry of Education categorizes TCM disciplines into three major groups:TCM disciplines‌ (including Chinese Medicine, Acupuncture and Tuina, Pediatric TCM, Orthopedics and Traumatology of TCM, etc.), Integrated Chinese and Western Medicine‌, Chinese Materia Medica disciplines‌ (including Chinese Pharmacy, Pharmaceutical Preparation of Chinese Medicine, Development of Chinese Medicinal Resources, etc.).

The Catalog of Disciplines and Specialties for Conferring Doctoral and Master's Degrees (1997) issued by the Ministry of Education classifies TCM disciplines into ‌Chinese Medicine‌, ‌Integrated Chinese and Western Medicine‌, and ‌Chinese Materia Medica‌. Among them, ‌Chinese Medicine‌ is subdivided into 13 specialties: Basic Theory of TCM, Clinical Foundations of TCM, Medical History and Literature of TCM, Chinese Medicinal Formulas, Diagnostics of TCM, Internal Medicine of TCM, Surgery of TCM, Orthopedics and Traumatology of TCM, Gynecology of TCM, Pediatrics of TCM, Ophthalmology \& Otorhinolaryngology of TCM, Acupuncture and Tuina, and Ethnic Medicine.

The Catalog of Medical Institution Clinical Departments (2007) issued by the Ministry of Health divides TCM clinical specialties into 18 categories: Internal Medicine, Surgery, Obstetrics \& Gynecology, Pediatrics, Dermatology, Ophthalmology, Otorhinolaryngology, Stomatology, Oncology, Orthopedics and Traumatology, Proctology, Geriatrics, Acupuncture, Tuina, Rehabilitation Medicine, Emergency Medicine, Preventive Healthcare, and others.

Based on the above classifications in TCM education and clinical practice, this study consolidates clinical specialties into ‌"‌Clinical Decision-making"‌, retains ‌Chinese Materia Medica‌ as a first-level discipline, and categorizes other sub-disciplines into ‌Classical Literature‌, ‌Acupuncture and Tuina‌, and ‌Basic Theory‌ for conciseness, professionalism, and interpretability. Thus, TCM-5CEVAL's dataset is classified into the following five categories.

\textbf{(1)‌Core Knowledge‌:} Denotes the scientific knowledge system encompassing fundamental concepts, principles, and theoretical frameworks of TCM.

\textbf{(2)‌Classical Literacy‌:} Comprises the TCM discipline that studies representative classical medical literature as primary research objects.

\textbf{(3)‌Clinical Decision-making‌:} Refers to the knowledge system for diagnosis, treatment, and prevention in clinical disciplines including internal medicine, surgery, gynecology, pediatrics, etc. of Traditional Chinese Medicine.

\textbf{(4)‌Chinese Materia Medica‌:} The discipline investigating the basic theories of Chinese medicines, including the origin, processing, properties, therapeutic effects, and clinical applications of medicinal materials, decoction pieces, and patent drugs.

\textbf{(5)‌Clinical Non-pharmacological Therapy‌:} The TCM discipline focusing on meridians, acupoints, and manual techniques, exploring operational skills, therapeutic principles, mechanisms of action, and disease prevention/treatment patterns.

\subsection{Evaluation dataset construction}
Building upon our previous TCM-3CEval framework, which established a triaxial benchmark for Core Knowledge, Classical Literacy, and Clinical Decision-Making, the TCM-5CEVAL dataset represents a significant methodological expansion. This new iteration retains the three original core dimensions while incorporating two critical, previously underserved domains: Chinese Materia Medica and Clinical Non-pharmacological Therapy, thus forming the expanded 5C evaluation architecture. The dataset is also rigorously constructed from authoritative data sources, primarily the "13th Five-Year Plan" and "14th Five-Year Plan" national planning textbook exercise sets. All content was curated and validated by subject-matter experts from Shanghai University of Traditional Chinese Medicine and the China Academy of Chinese Medical Sciences to ensure accuracy and clinical relevance.

The TCM-5CEVAL dataset is composed of five sub-datasets, corresponding to each of the 5C dimensions. To facilitate a multi-faceted evaluation that assesses both knowledge recall and complex reasoning, each sub-dataset contains a collection of single-choice questions, multiple-choice questions, and open-ended questions. This design moves beyond solely objective metrics to capture a model's capacity for subjective judgment and in-depth analysis. Furthermore, the selection of questions was stratified to ensure a balanced distribution of difficulty, encompassing easy, medium, and hard items. The specific distribution of question difficulty across dimensions is detailed in Table \ref{tab:5c_dimensions_detailed}. The five sub-datasets are defined as follows:

1. \textbf{TCM-Exam (Core Knowledge)}: This dataset evaluates the model's foundational understanding of TCM. It assesses the comprehension of core theoretical constructs (e.g., Yin-Yang, Five Elements, Zang-Fu, Qi-Blood-Body Fluids) and the application of fundamental diagnostic principles, including the four diagnostic methods and various syndrome differentiation systems.

2. \textbf{TCM-LitQA (Classical Literacy)}: This dataset measures the model's proficiency in interpreting seminal TCM literature. It features questions derived from the four major classics (Huangdi Neijing, Shanghan Lun, Jingui Yaolue, Wenbing Xue) and the theories of various TCM schools. The assessment focuses on the model's ability to analyze classical provisions and understand their enduring theoretical and clinical implications.

3. \textbf{TCM-MRCD (Clinical Decision-Making)}: This dataset assesses the model's capacity for practical clinical reasoning. Using standardized clinical case records, it evaluates the entire diagnostic and therapeutic process: from analyzing patient data and performing syndrome differentiation to formulating treatment principles and appropriate prescriptions.

4. \textbf{TCM-CMM (Chinese Materia Medica)}: This new dimension evaluates the model's specialized knowledge of herbal medicine and formulary. It covers the properties, efficacy, and clinical application of individual herbs, as well as the principles of formula composition, compatibility (including contraindications), processing (Paozhi), and quality assessment.

5. \textbf{TCM-ClinNPT (Clinical Non-pharmacological Therapy)}: This second new dimension addresses non-pharmacological interventions, focusing on acupuncture, moxibustion, and Tuina. The dataset tests the model's ability to perform syndrome differentiation for these therapies, select appropriate acupoints, design Tuina manipulation protocols, and apply these skills in common clinical scenarios.

\begin{table}[ht]
\centering
\caption{TCM-5CEval Benchmark Dimensions and Question Distribution}
\label{tab:5c_dimensions_detailed}
\begin{tabular}{@{}llccp{8cm}@{}}
\toprule
\textbf{Dimension} & \textbf{Question Type} & \textbf{N} & \textbf{Difficulty (E/M/H)} & \textbf{Main Evaluation Content} \\
\midrule

\multirow{3}{*}{\textbf{TCM-Exam}} 
 & Single-choice & 122 & 38 / 52 / 32 & Assess recall of core concepts, theories, and diagnostic facts. \\
 & Multiple-choice & 78 & 29 / 29 / 20 & Evaluate understanding of complex relationships between foundational theories. \\
 & Open-ended & 70 & 22 / 32 / 16 & Examine in-depth explanation of theoretical principles and diagnostic logic. \\
\cmidrule(l){2-5}

\multirow{3}{*}{\textbf{TCM-LitQA}} 
 & Single-choice & 275 & 97 / 96 / 82 & Test comprehension of key provisions and ideas from classical texts. \\
 & Multiple-choice & 188 & 61 / 71 / 56 & Assess ability to compare and contrast concepts across different classics. \\
 & Open-ended & 165 & 52 / 64 / 49 & Evaluate deep interpretation of classical texts and their clinical significance. \\
\cmidrule(l){2-5}

\multirow{3}{*}{\textbf{TCM-MRCD}} 
 & Single-choice & 230 & 80 / 98 / 52 & Evaluate diagnostic accuracy in straightforward clinical vignettes. \\
 & Multiple-choice & 156 & 59 / 53 / 44 & Test syndrome differentiation and treatment selection in complex cases. \\
 & Open-ended & 148 & 45 / 59 / 44 & Assess the complete clinical reasoning process (diagnosis, principle, formula). \\
\cmidrule(l){2-5}

\multirow{3}{*}{\textbf{TCM-CMM}} 
 & Single-choice & 273 & 110 / 100 / 63 & Assess recall of herb properties, formula compositions, and processing facts. \\
 & Multiple-choice & 121 & 43 / 42 / 36 & Evaluate understanding of herb compatibility, contraindications, and formula logic. \\
 & Open-ended & 91 & 33 / 31 / 27 & Examine the ability to analyze formulas and compare/contrast materia medica. \\
\cmidrule(l){2-5}

\multirow{3}{*}{\textbf{TCM-ClinNPT}} 
 & Single-choice & 166 & 55 / 64 / 47 & Test knowledge of acupoint locations, functions, and Tuina techniques. \\
 & Multiple-choice & 96 & 38 / 30 / 28 & Assess acupoint/technique selection for specific clinical conditions. \\
 & Open-ended & 81 & 25 / 32 / 24 & Evaluate the design of complete acupuncture or Tuina treatment plans. \\
\bottomrule
\end{tabular}
\end{table}

\subsection{Workflow and Evaluation Methods of TCM-5CEVAL}
\begin{figure}
    \centering
    \includegraphics[width=1\linewidth]{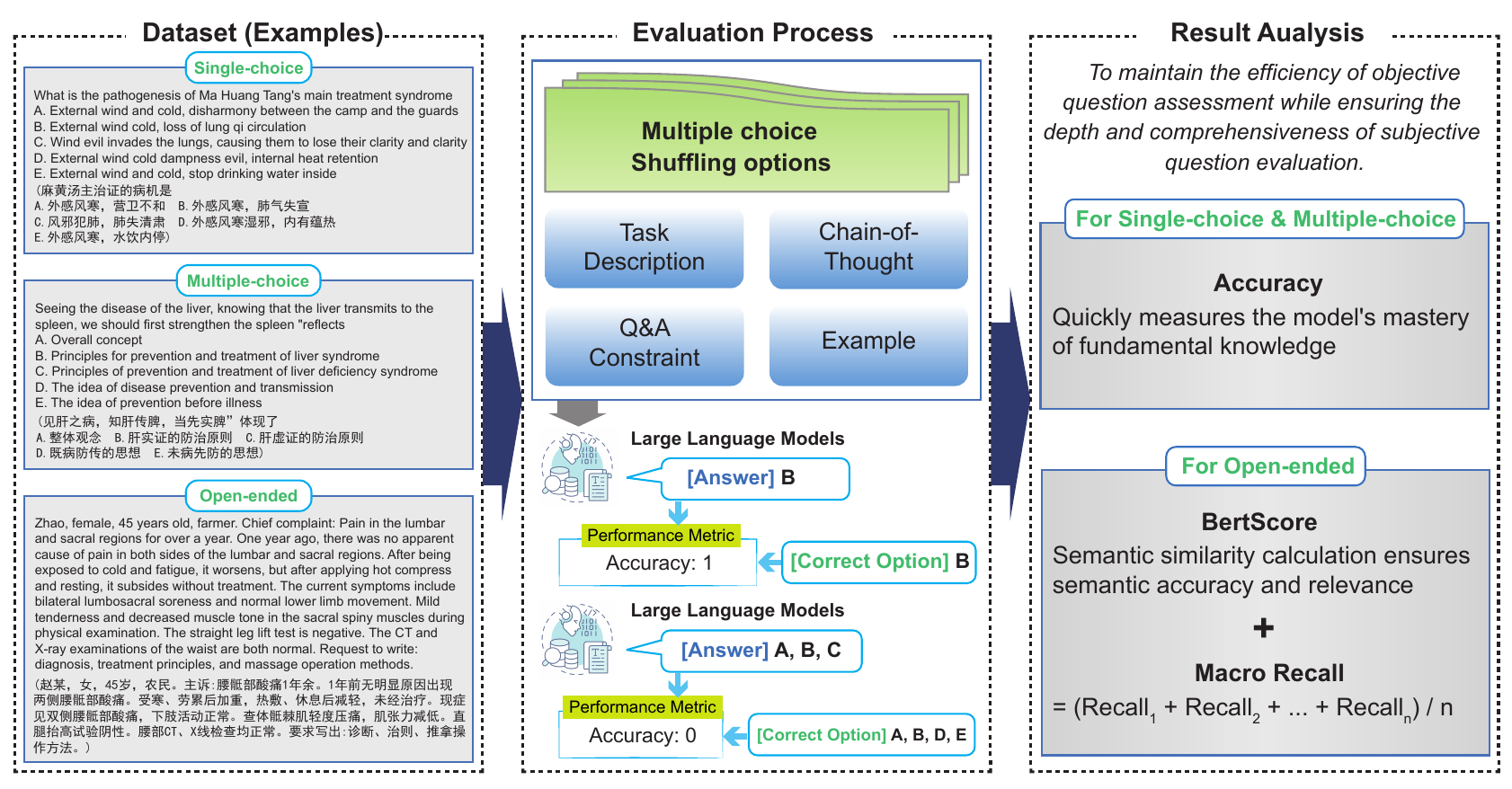}
    \caption{TCM-LLM Multi-Metric Assessment Workflow}
    \label{fig:workflow}
\end{figure}

The proposed TCM-LLM Multi-Metric Assessment Workflow(\autoref{fig:workflow}) systematically evaluates model performance through a dual-path framework that employs accuracy metrics for objective questions (single-choice and multiple-choice questions) and combines BertScore with macro recall for open-ended responses.

To further assess the stability of the answers and the robustness against option-order bias, we specifically applied a permutation-based consistency test for the single-choice questions. Each original single-choice question contains five mutually exclusive options (A–E). For each question we generate five permutations of the option ordering by cyclically rotating the original option sequence so that each original option appears in each position exactly once. Each model is presented with all five permutations and produces one prediction per permutation. A question is considered passed by a given model only if the model outputs the same correct option for all five permutations (i.e., the model’s answers across permutations are identical and equal to the annotated correct choice). This strict consistency criterion evaluates whether model predictions are invariant to option ordering and reduces the influence of positional biases. To ensure the evaluation results reflect only the model's robustness against option ordering rather than the randomness of the generation process, all model inferences were conducted with the temperature set to $0$, guaranteeing deterministic outputs.

\section{Results}
The performance of fifteen large language models was evaluated across the five dimensions of the TCM-5CEVAL benchmark. The results, presented in Table \ref{tab:tcm_benchmarks}, indicate a wide range of capabilities among the models in comprehending and applying Traditional Chinese Medicine knowledge.

\subsection{Overall Model Performance Hierarchy}
A clear performance hierarchy was observed among the evaluated models. Kimi\_K2\_Instruct\_0905 demonstrated strong overall performance, achieving the highest scores in three of the five sub-datasets: TCM-Exam (0.847), TCM-MRCD (0.746), and TCM-CMM (0.749). Concurrently, deepseek\_r1 also consistently achieved high scores, securing the top rank in TCM-LitQA (0.731) and ranking second in TCM-Exam (0.798) and TCM-CMM (0.746). Other models, including gemini\_2\_5\_pro and grok\_4\_0709, also constituted a high-performing group, with scores exceeding 0.700 in multiple dimensions.

\subsection{Performance Variation Across 5C Dimensions}
A consistent pattern observed across most models was a performance differential among the five evaluation dimensions. The highest scores were typically registered on the TCM-Exam dataset, which assesses core knowledge. For instance, Kimi\_K2\_Instruct\_0905 achieved its highest score of 0.847 on TCM-Exam. In contrast, the TCM-LitQA dimension, focusing on classical literacy, appeared more challenging for many models. Several models that performed well on core knowledge exhibited a drop in performance on this sub-dataset; for example, gpt\_5 scored 0.750 on TCM-Exam but 0.519 on TCM-LitQA. A similar trend of lower relative scores was also apparent in the TCM-ClinNPT dimension for some models.

\begin{table}[htbp]
\centering
\caption{Comparison of performance of different models on the TCM-5CEVAL benchmark}
\label{tab:tcm_benchmarks}
\renewcommand{\arraystretch}{1.3}
\begin{tabular}{l ccccc}
\toprule

\textbf{Model} & \textbf{TCM-Exam} & \textbf{TCM-LitQA} & \textbf{TCM-MRCD} & \textbf{TCM-CMM} & \textbf{TCM-ClinNPT} \\
\midrule

deepseek\_r1 & \cellcolor{lightblue}0.798 & \cellcolor{lightred}\best{0.731} & \cellcolor{lightblue}0.733 & \cellcolor{lightblue}0.746 & \cellcolor{lightblue}0.640 \\
deepseek\_v3 & 0.770 & 0.611 & 0.706 & 0.659 & 0.565 \\
qwen3\_235b & 0.764 & 0.626 & 0.691 & 0.704 & 0.566 \\
qwen3\_32b & 0.724 & 0.549 & 0.657 & 0.607 & 0.508 \\
qwen2\_5\_72b & 0.754 & 0.550 & 0.660 & 0.683 & 0.525 \\
\hdashline[4pt/4pt]

Kimi\_K2\_Instruct\_0905 & \cellcolor{lightred}\best{0.847} & \cellcolor{lightblue}0.696 & \cellcolor{lightred}\best{0.746} & \cellcolor{lightred}\best{0.749} & 0.595 \\
gemini\_2\_5\_flash & 0.675 & 0.492 & 0.635 & 0.536 & 0.493 \\
gemini\_2\_5\_pro & 0.779 & 0.62 & 0.724 & 0.726 & 0.612 \\
gpt\_4o & 0.665 & 0.469 & 0.609 & 0.571 & 0.479 \\
gpt\_5 & 0.750 & 0.519 & 0.641 & 0.666 & 0.606 \\

\hdashline[4pt/4pt]
gpt\_5\_mini & 0.629 & 0.425 & 0.623 & 0.547 & 0.513 \\
mistral\_small\_3\_1\_24b\_instruct & 0.454 & 0.347 & 0.441 & 0.39 & 0.368 \\
llama\_4\_maverick & 0.721 & 0.513 & 0.638 & 0.544 & 0.443 \\
grok\_4\_0709 & 0.730 & 0.593 & 0.684 & 0.680 & \cellcolor{lightred}\best{0.642} \\
claude\_sonnet\_4\_5\_20250929 & 0.698 & 0.593 & 0.672 & 0.717 & 0.560 \\
\bottomrule
\end{tabular}

\vspace{0.5em}
{\footnotesize 
\colorboxrule{lightred}{\ } marks the best score in a column and  \colorboxrule{lightblue}{\ }the second best.
}

\end{table}

\subsection{Intra-Family Model Comparisons}
The results also highlight performance variations within specific model families, often correlating with model size or version. For instance, deepseek\_r1 consistently outperformed deepseek\_v3 across all five categories. A similar trend was observed for the Gemini series, where gemini\_2\_5\_pro achieved significantly higher scores than gemini\_2\_5\_flash. Within the Qwen family, the largest model, qwen3\_235b, generally surpassed its smaller counterparts, qwen3\_32b and qwen2\_5\_72b. Likewise, gpt\_5 recorded higher scores than both gpt\_4o and gpt\_5\_mini, with gpt\_5\_mini showing the lowest performance within its family.

\subsection{Competency in Materia Medica and Non-Pharmacological Therapies}
The two novel dimensions, TCM-CMM and TCM-ClinNPT (Clinical Non-pharmacological Therapy), introduced to assess specialized and practical knowledge, presented varied results.In the TCM-CMM dataset, which assesses knowledge of herbal medicine and formulary, Kimi\_K2\_Instruct\_0905 obtained the highest score (0.749), followed closely by deepseek\_r1 (0.746). In the domain of non-pharmacological therapies (TCM-ClinNPT), grok\_4\_0709 registered the top score at 0.642, demonstrating a particular capability in knowledge related to acupuncture, moxibustion, and Tuina. The second-highest score in this category was achieved by deepseek\_r1 (0.640).

\subsection{Robustness to Option-Order Permutation}
\begin{figure}
    \centering
    \includegraphics[width=1.0\linewidth]{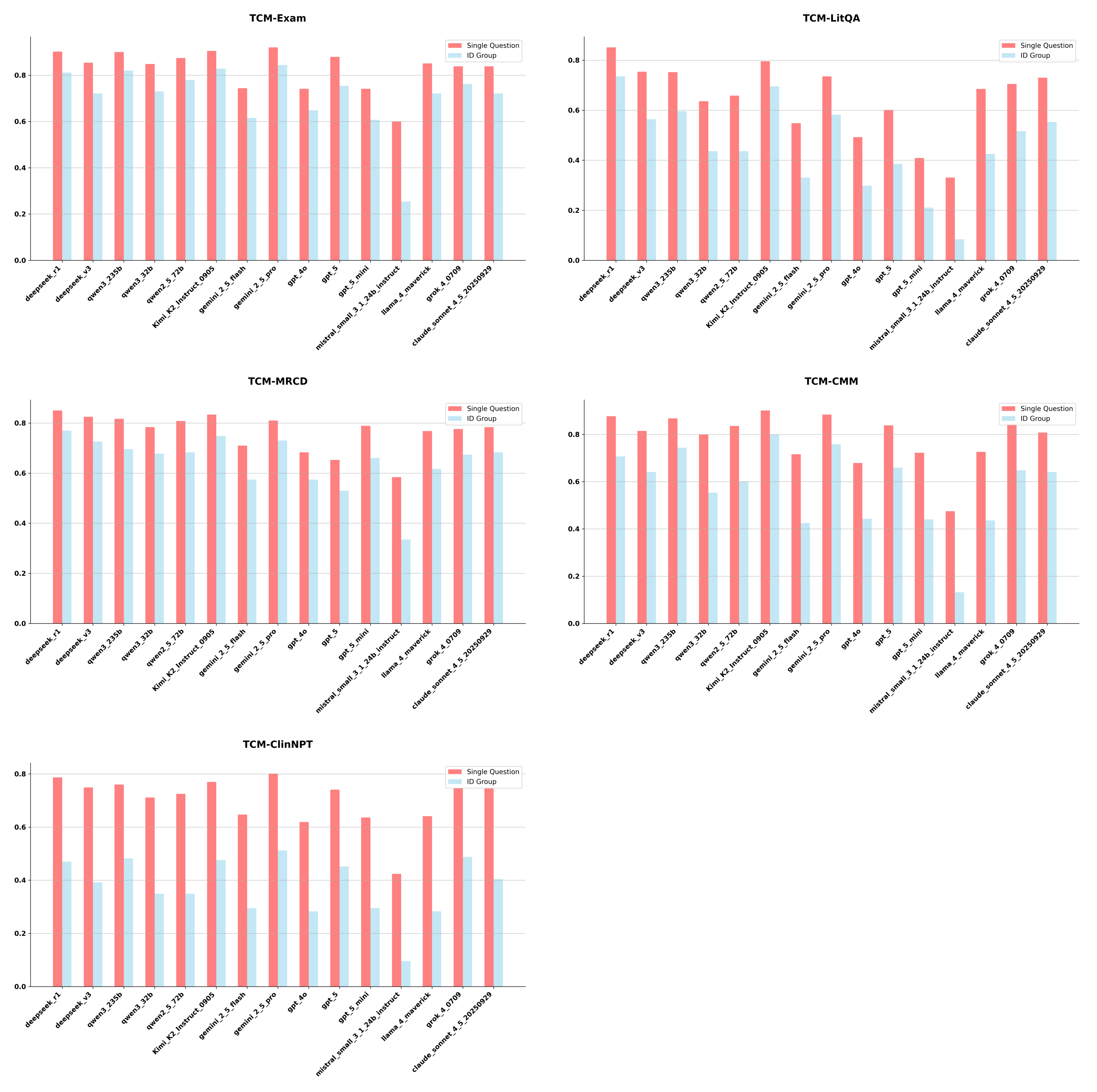}
    \caption{Model Performance on the Single-Choice Permutation Consistency Test. The Figure compares the standard accuracy on the original questions ('Single Question') with the consistency-based accuracy ('ID Group') for each model across the five sub-datasets. The 'ID Group' score is awarded only when a model correctly answers a question across all five cyclical permutations of its options, thus measuring its robustness against option-order bias.}
    \label{fig:5C Robustness}
\end{figure}

To further probe the stability of model predictions, a permutation-based consistency test was applied to the single-choice questions, where models were required to correctly identify the answer across five different orderings of the options. The results, detailed in Figure \ref{fig:5C Robustness}, indicate a universal performance degradation for all models across all five sub-datasets when this strict consistency criterion was enforced, compared to their baseline accuracy on single-question instances. The magnitude of this performance drop varied among models and datasets. For instance, on the TCM-Exam sub-dataset, gemini\_2\_5\_pro demonstrated relatively high consistency, with its accuracy decreasing moderately from a baseline of 0.920 to a consistency-tested score of 0.844. However, the performance degradation was substantially more pronounced on other sub-datasets, indicating greater sensitivity to option ordering. On the TCM-LitQA sub-dataset, the accuracy of gpt\_4o fell from 0.492 to 0.298, and mistral\_small\_3\_1\_24b\_instruct saw its score drop from 0.331 to 0.084. The effect was even more severe on the TCM-ClinNPT sub-dataset, where deepseek\_r1’s score declined from 0.787 to 0.470. This trend of reduced accuracy under the permutation test was consistently observed, though the degree of reduction differed across the evaluated models and the specific knowledge dimension being tested.

\section{Discussion}
Our prior TCM-3CEval framework provided an important foundational step, establishing a triaxial benchmark for Core Knowledge, Classical Literacy, and Clinical Decision-Making. It serves as a useful tool for assessing a model's grasp of foundational theory and its diagnostic reasoning processes. A limitation of the 3C framework, however, was that its "Clinical Decision-Making" dimension consolidated all therapeutic recommendations. This design did not provide a granular assessment that discretely evaluated the two primary interventional arms of TCM: pharmacological therapy (herbal medicine)\cite{lee2022systems} and non-pharmacological therapy (acupuncture, Tuina)\cite{tan2024protocol}.

The TCM-5CEval framework builds upon this foundation to offer enhanced specificity and breadth. The first key change was the formal decoupling of clinical therapeutics. We expanded the 3C framework's clinical assessment by introducing two new, dedicated dimensions: Chinese Materia Medica (TCM-CMM) and Clinical Non-pharmacological Therapy (TCM-ClinNPT). This architectural change enables a more detailed and specialized evaluation of a model's distinct competencies in each of these therapeutic modalities. The second main enhancement lies within the new TCM-CMM dimension, which represents an expansion in scope. The 3C benchmark's assessment of herbal medicine was largely confined to formula composition and modification within a clinical context. The 5C framework's TCM-CMM dimension is more comprehensive. It assesses a model's knowledge not only in clinical formula application but also in the foundational pharmaceutical science of TCM. This includes dedicated evaluations of pharmacological analysis\cite{wang2022traditional}, traditional herb processing (Paozhi)\cite{luo2021key}, pharmaceutics\cite{xu2021comprehensive}, and herb identification/quality control\cite{balekundri2020quality}. This expansion marks an important shift. The benchmark's utility is no longer limited to assessing clinical acumen; it now systematically evaluates a model's capacity in areas relevant to TCM-related scientific research\cite{chu2020quantitative}, pharmaceutical development\cite{li2025advancing}, and quality assurance\cite{zhang2022research}. This makes TCM-5CEval a more complete instrument for measuring a model's proficiency across a wider spectrum of the modern TCM domain.

Our evaluation of fifteen large language models on the TCM-5CEVAL benchmark revealed significant performance disparities, as detailed in Table~\ref{tab:tcm_benchmarks}. A clear performance hierarchy was observed. Models such as Kimi\_K2\_Instruct\_0905, deepseek\_r1, and gemini\_2\_5\_pro formed a high-performing group, achieving top scores across multiple dimensions. For instance, Kimi\_K2\_Instruct\_0905 scored highest overall in three of the five primary dimensions: TCM-Exam, TCM-MRCD, and TCM-CMM. deepseek\_r1 led in TCM-LitQA, while grok\_4\_0709 secured the top score in TCM-ClinNPT. A consistent pattern emerged across most models: performance varied significantly across the 5C dimensions; scores were typically highest on TCM-Exam, which assesses core knowledge, but dropped notably on TCM-LitQA, which requires interpretative analysis of classical texts. This suggests a general proficiency in recalling foundational concepts but a widespread difficulty with deeper interpretative and reasoning tasks. Furthermore, the permutation-based consistency test (Figure~\ref{fig:5C Robustness}) demonstrated that this knowledge mastery is not robust; all models exhibited a universal performance degradation when faced with varied option ordering, indicating a sensitivity to positional biases and a lack of deep, stable understanding.

\begin{figure}
    \centering
    \includegraphics[width=1\linewidth]{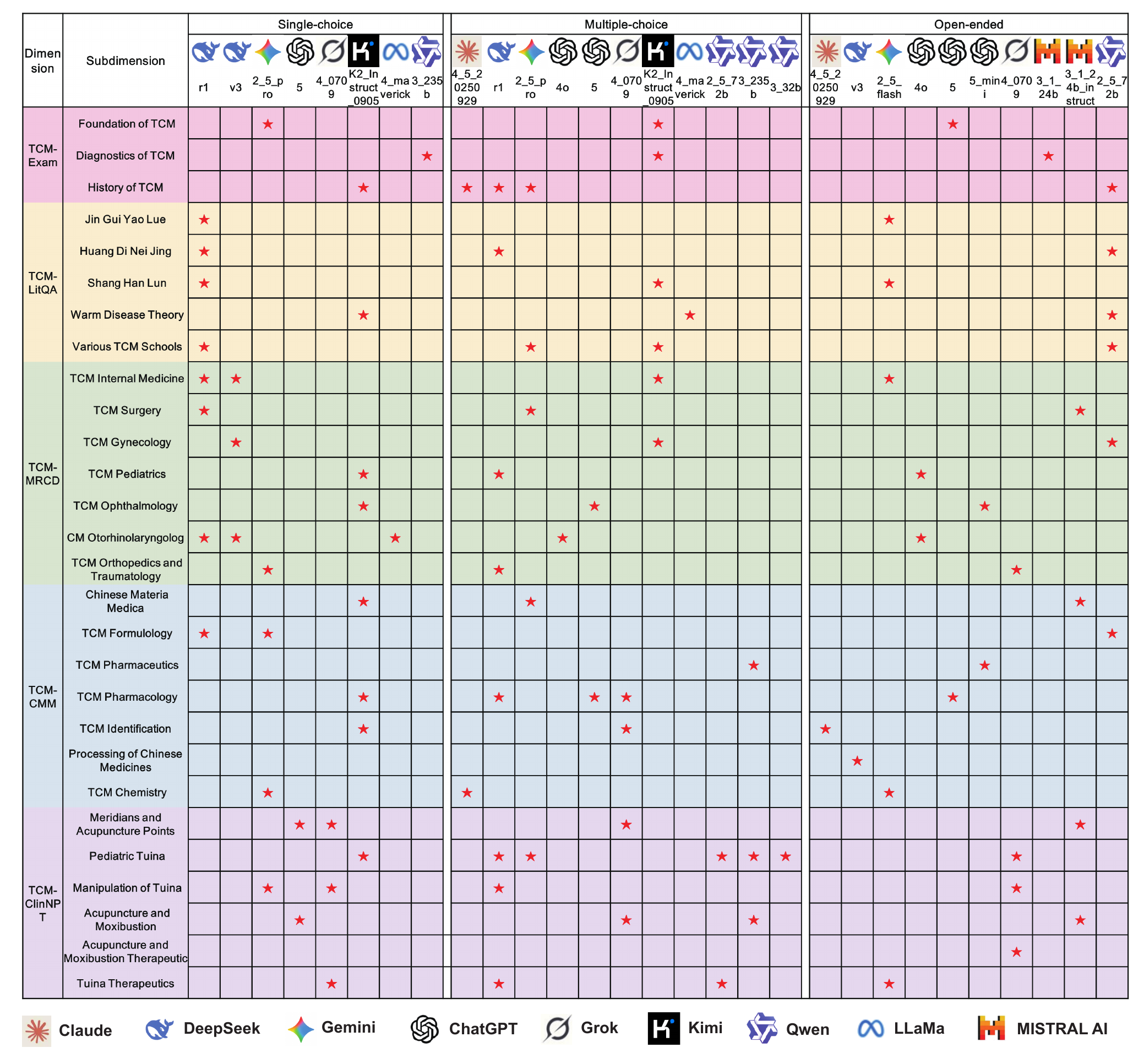}
    \caption{Performance Distribution of Leading Models by Sub-Dimension. \ding{72}: The model that performs best in this sub-dimension}.
    \label{top1}
\end{figure}

To further dissect these high-level trends, we conducted a two-part investigation. First, we performed a granular comparison of the top-performing models at the sub-dimensional level to identify areas of specialization. Second, to identify common weaknesses, we conducted an error hotspot analysis by aggregating questions with universally poor performance. These were defined as objective items (single and multiple-choice) answered incorrectly by five or more models, and subjective items (open-ended) where the average model score ranked in the lowest quartile (bottom 25\%) . Our analysis of specialized strengths reveals that top-performing models exhibit distinct and specialized competencies, suggesting their capabilities in the vertical domain of TCM are not uniform. For example, deepseek\_r1 demonstrated a profound specialization in classical literacy and non-pharmacological therapies; it dominated the single-choice questions for TCM-LitQA (winning 4 of 7 sub-categories, including Shanghan Lun and Neijing) and swept all 4 sub-categories of the TCM-ClinNPT multiple-choice questions (e.g., Tuina techniques and Acupuncture). In contrast, Kimi\_K2\_Instruct\_0905 showed broad strength in foundational knowledge and modern pharmaceutical applications, securing top performance in TCM-Exam multiple-choice questions and multiple TCM-CMM single-choice sub-categories (Pharmaceutics, Pharmacology). gemini\_2\_5\_pro distinguished itself in highly technical areas, such as Chinese Materia Medica Chemistry and Processing (Paozhi). These specializations likely reflect significant differences in the composition of the models' respective training corpora. A model's dominance in classical texts (deepseek\_r1) implies a corpus heavily enriched with digitized classical literature, whereas strength in materia medica chemistry (gemini\_2\_5\_pro) or pharmacology (Kimi\_K2\_Instruct\_0905) suggests a focus on modern textbooks, scientific papers, and pharmaceutical data. This divergence indicates that achieving expert-level performance across the entire breadth of TCM may require highly specialized, curated datasets rather than general-purpose training. The specific distribution of top-performing models across individual sub-dimensions is presented in the Figure\ref{top1}.

\begin{figure}
    \centering
    \includegraphics[width=1\linewidth]{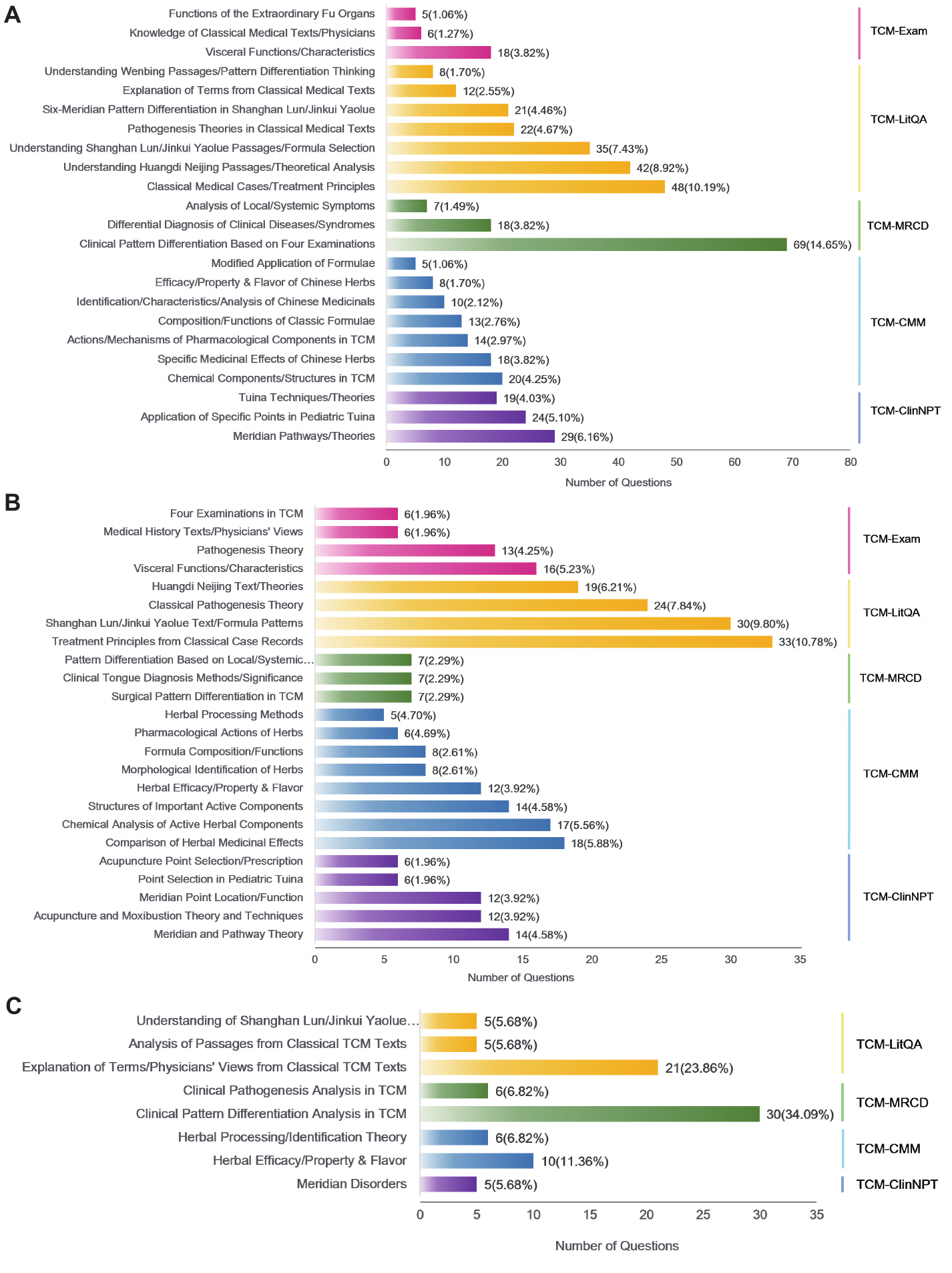}
    \caption{High-Frequency Errors in Question Sets (A. Single-Choice; B. Multiple-Choice; C. Open-Ended)}
    \label{error}
\end{figure}

This divergence in model strengths, which points to training data biases, is counterbalanced by a consistent and universal set of weaknesses revealed by our error hotspot analysis. These failures manifested differently across question types but consistently pointed to the same core inferential deficits. On objective questions, while models showed some recall, their performance was not robust. A specific analysis of the aggregated single-choice errors shows the largest cluster was `clinical four-diagnosis pattern differentiation' (69 instances), followed by difficulties in classical text interpretation, such as `classical case studies' (48 instances).As shown in the Figure\ref{error}(A), these are the error-prone knowledge points for single-choice questions.

This pattern was mirrored in the aggregated multiple-choice question errors. Here, models struggled with comparative analysis---such as differentiating the similarities and differences in drug efficacies---and again, showed weakness in understanding classical case studies and original texts from Shanghan Lun and Jingui Yaolue.The Figure\ref{error}(B) illustrates the most commonly mistaken knowledge points in multiple-choice questions.

Conversely, on subjective (open-ended) questions, the analysis of low-scoring items shows the most frequent challenge, by a large margin, was `TCM clinical pattern differentiation analysis' (30 instances). This was followed by difficulties in explaining classical literature terminology.The distribution of commonly misinterpreted concepts in constructed-response questions is visualized in the Figure\ref{error}(C).

This convergence of errors across all three question types points to several key weaknesses: (1) a noted inability to perform textual exegesis on classical literature; (2) a significant deficit in practical Zheng (syndrome) differentiation, which was a top error in both objective and subjective formats; and (3) difficulty with nuanced, expert-level knowledge like materia medica comparisons. These deficiencies likely stem from two primary causes: data sparsity and linguistic barriers related to the archaic, high-context language of TCM classics; and a more fundamental inferential deficit. This explains why models fail at both robust objective reasoning and generative subjective synthesis, as they have not yet replicated the holistic, dialectical reasoning (`treatment based on syndrome differentiation') that is central to TCM practice.

While this 5C framework offers a more comprehensive evaluation, we identify several directions for future work. The current datasets are primarily derived from authoritative textbooks and literature cases. A critical next step involves validating model performance against large-scale, real-world clinical data, such as electronic medical records, to ensure practical applicability. Furthermore, the challenge of polysemous terminology\cite{satibaldieva2024polysemy, hou2025pruning, song2025status} in TCM remains ; future work may explore concept disambiguation mechanisms\cite{wang2020word}, perhaps by integrating expert annotations\cite{levy2021assessing} with structured medical knowledge graphs\cite{li2020real, gao2025leveraging}. Finally, this benchmark is text-based. A long-term goal is to expand the framework to include multi-modal evaluation, which is essential for assessing a model's capacity to interpret non-textual diagnostic information central to TCM, such as tongue and pulse diagnosis.

\section{Conclution}
This study introduces TCM-5CEVAL, a comprehensive benchmark for evaluating large language models across five core dimensions of Traditional Chinese Medicine. Our evaluation of fifteen models revealed significant performance disparities, with models like deepseek\_r1, Kimi\_K2\_Instruct\_0905, and gemini\_2\_5\_pro demonstrating strong capabilities. A key finding is the uneven performance across domains, where models generally excelled at recalling foundational concepts (TCM-Exam) but were less proficient in the interpretative analysis of classical literature (TCM-LitQA). Critically, a permutation-based consistency test revealed a universal lack of robustness; all models, including top performers, exhibited a notable performance degradation when challenged with varied option ordering, indicating a sensitivity to positional biases. While leading LLMs show promise in the TCM domain, these findings underscore that their knowledge is inconsistent and their reasoning remains fragile. TCM-5CEVAL thus serves as a crucial tool for diagnosing these weaknesses and guiding future efforts toward developing more knowledgeable and fundamentally reliable models.

\section*{Acknowledgments}
This work was supported by the 2022 National Natural Science Foundation of
China [grant 82174506], the 2024 Traditional Chinese Medicine Research Project of Shanghai Municipal Health Commission [grant 2024PT001] and 2025 Traditional Chinese Medicine Standardization Project of Shanghai Administration of Traditional Chinese Medicine [grant 2025BZ002].

\section*{Data Availability Statement}
The datasets used in this study are available through the MedBench open platform at https://medbench.opencompass.org.cn/home. Access to the data can be obtained by contacting the MedBench team or the corresponding author.

\section*{Declarations}
The authors declare that the research was conducted in the absence of any commercial or financial relationships that could be construed as a potential conflict of interest.

\bibliographystyle{unsrt}  
\bibliography{references}  

\clearpage 
\appendix 

\section{MORE DETAILS OF TCM-5CEVAL}

\subsection{PROMPTS FOR TCM-5CEVAL}
The prompts used in the TCM-5CEval benchmark are standardized instructions, meticulously designed to guide Large Language Models (LLMs) toward producing outputs that are uniform and optimized for automated evaluation across different question modalities. Each prompt's design comprises two primary components:

\textbf{Role-Playing Instruction:} A consistent directive, "You are a TCM domain expert," is employed across all prompts. This primes the model to activate its specialized knowledge base in Traditional Chinese Medicine, thereby enhancing the accuracy and professional quality of its responses.

\textbf{Strict Formatting Constraints:} The prompts enforce a set of rigid output formats tailored to the respective question type. This is the cornerstone of the benchmark's automated scoring capability:

For single-choice questions, the model must return a single letter enclosed in angle brackets (e.g., <B>).

For multiple-choice questions, the model must provide all correct letters, comma-separated, within angle brackets (e.g., <ACD>).

For open-ended questions, the model is instructed to generate a direct, concise textual answer, stripped of any conversational filler or extraneous explanations.

This disciplined approach ensures that all outputs are machine-readable and structurally consistent, enabling efficient, reliable, and scalable evaluation across the entire benchmark.

\begin{figure}[H]
    \centering
    \includegraphics[width=1.0\linewidth]{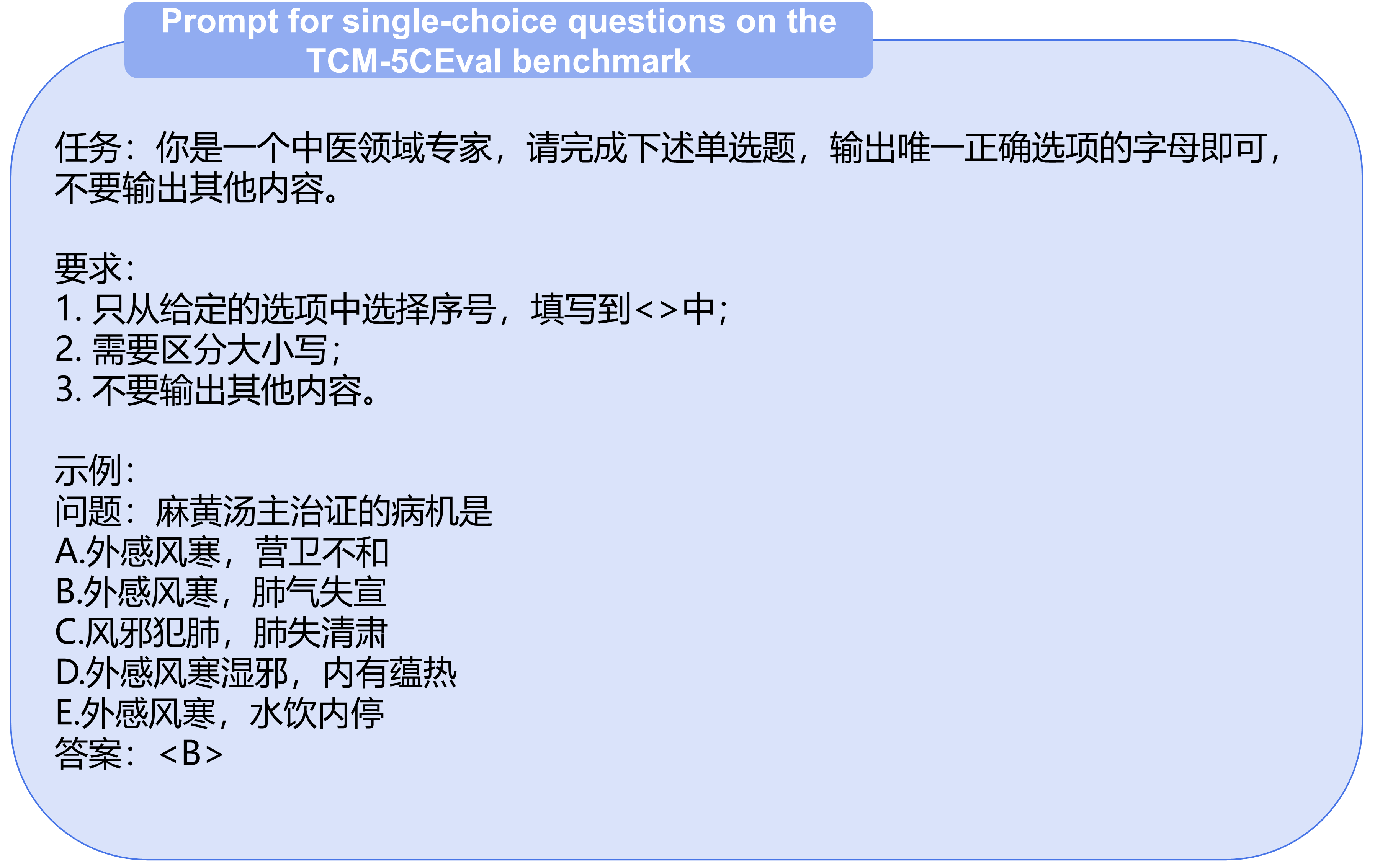}
    \caption{Prompt for single-choice questions on the TCM-5CEval benchmark}
    \label{fig:prompt-1}
\end{figure}

\begin{figure}
    \centering
    \includegraphics[width=1.0\linewidth]{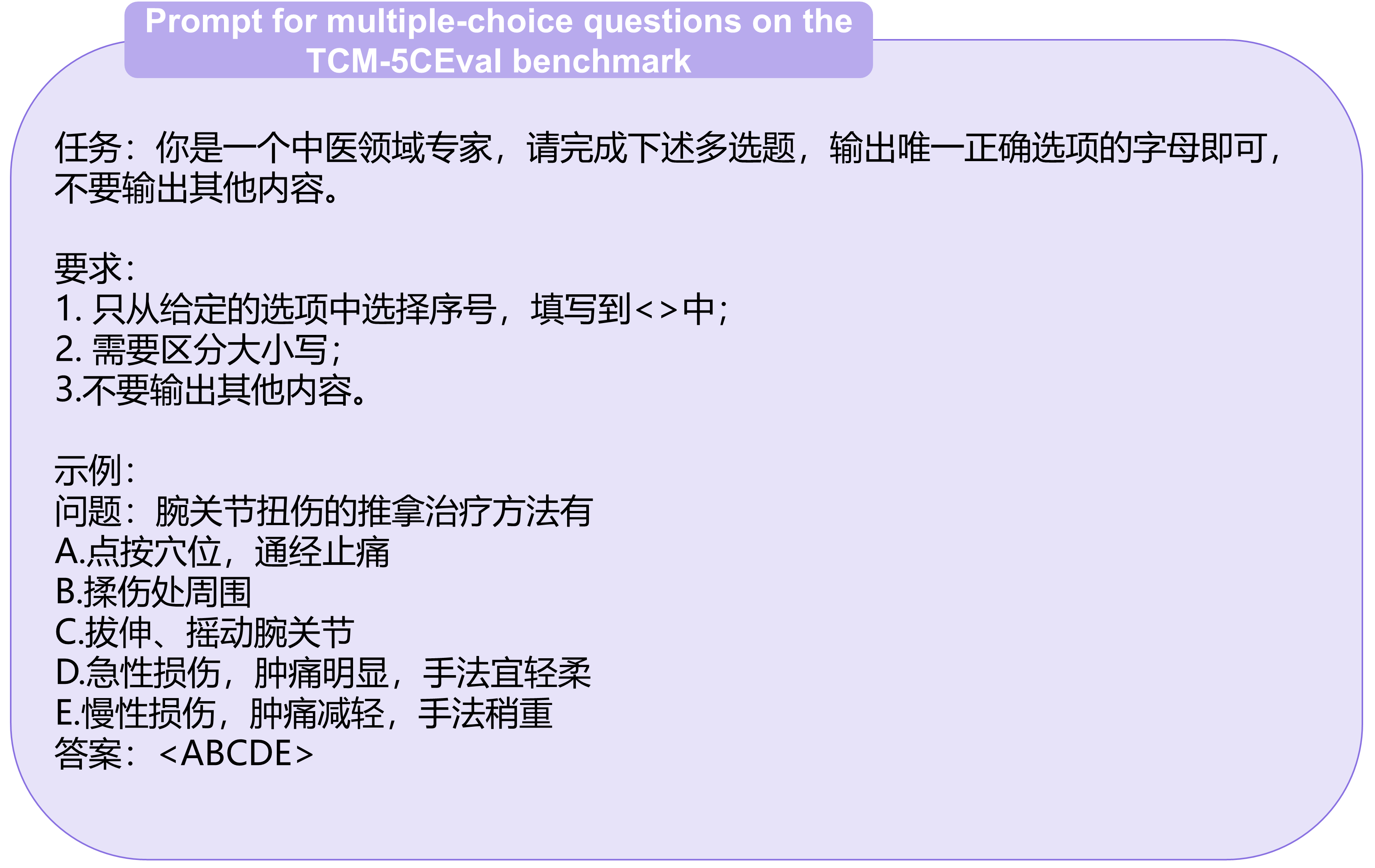}
    \caption{Prompt for multiple-choice questions on the TCM-5CEval benchmark}
    \label{fig:prompt-2}
\end{figure}

\begin{figure}
    \centering
    \includegraphics[width=1.0\linewidth]{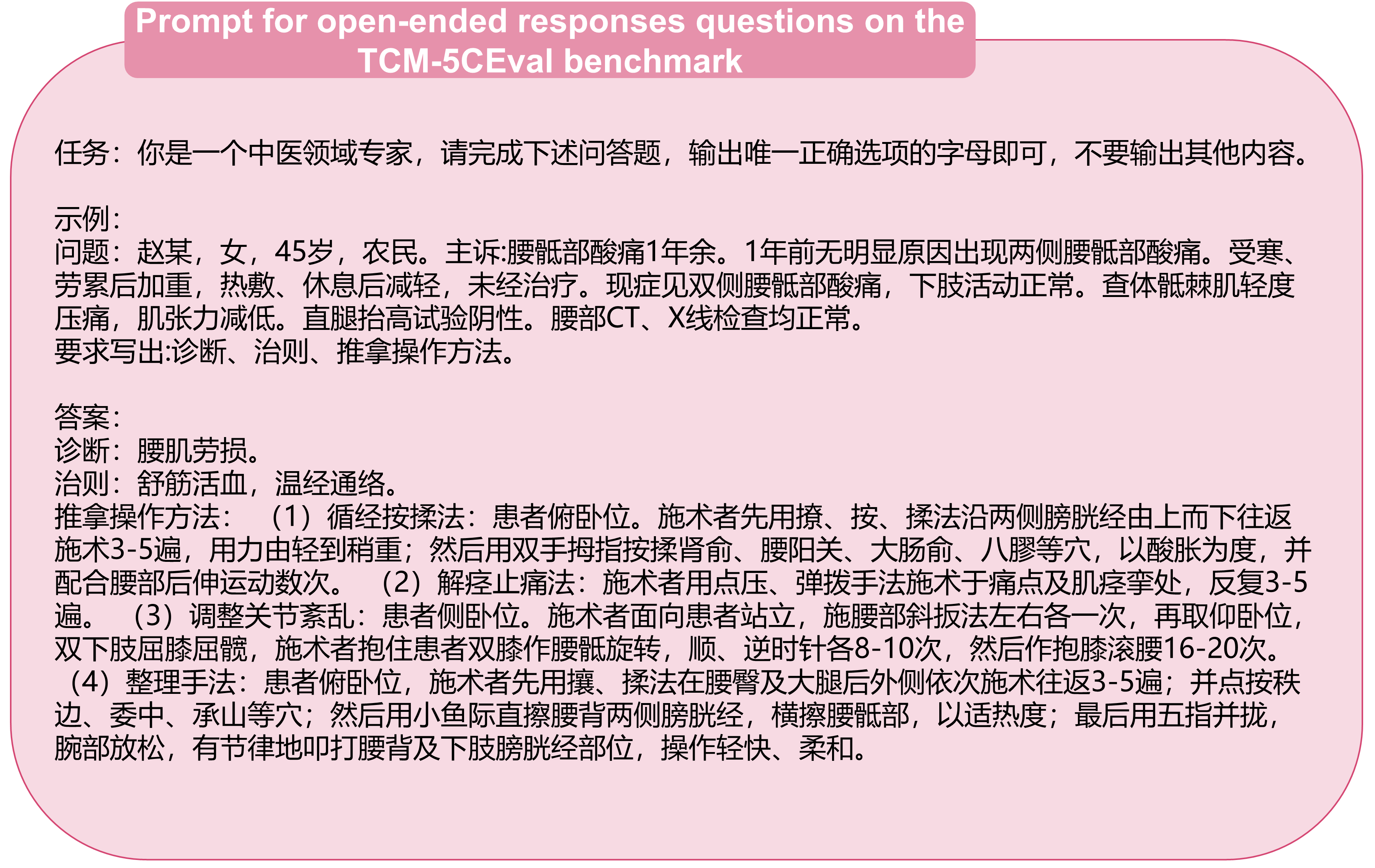}
    \caption{Prompt for open-ended responses questions on the TCM-5CEval benchmark}
    \label{fig:prompt-3}
\end{figure}

\end{document}